\pdfoutput=1

\documentclass[11pt]{article}

\usepackage{acl}

\usepackage{times}
\usepackage{latexsym}

\usepackage[T1]{fontenc}

\usepackage[utf8]{inputenc}

\usepackage{microtype}

\usepackage{inconsolata}

\usepackage{xurl}

\usepackage{amsmath}
\usepackage{stmaryrd}
\usepackage{verbatim}

\usepackage{tabularx}               %
\usepackage{booktabs}
\newcolumntype{C}{>{\centering\arraybackslash}X}
\usepackage{multirow}               %
\usepackage{diagbox}                %
\usepackage{hhline}                 %
\usepackage{color}                  %
\usepackage{amsmath}                %
\usepackage{amssymb}                %
\usepackage{mathtools}              %
\usepackage{soul}                   %

\definecolor{RoseQuartzBg}{HTML}{F7CAC9}
\definecolor{RoseQuartz}{HTML}{F5A798}
\definecolor{Serenity}{HTML}{92A8D1}
\definecolor{OrangeRed}{rgb}{1.0, 0.27, 0.0}
\definecolor{Turquoise}{HTML}{0F4C81}

\usepackage{enumitem}

\usepackage{arydshln}               %
\definecolor{themered}{HTML}{FF8375}

\usepackage{xparse}
\usepackage{bbm}
\usepackage{float}
\usepackage{caption}
\usepackage{subcaption}
\usepackage{stfloats}
\usepackage{pifont}
\newcommand*\colourcheck[1]{%
  \expandafter\newcommand\csname #1check\endcsname{\textcolor{#1}{\ding{52}}}%
}
\newcommand*\colourxmark[1]{%
  \expandafter\newcommand\csname #1xmark\endcsname{\textcolor{#1}{\ding{56}}}%
}

\colourcheck{green}
\colourxmark{red}
\usepackage{cancel}
\usepackage{transparent}

\NewDocumentCommand{\haoyang}{ mO{} }{\textcolor{Turquoise}{\textsuperscript{\textsc{Haoyang}}\textsf{\textbf{\small[#1]}}}}

\DeclareMathOperator{\enc}{\textit{Enc}}
\usepackage{dutchcal}

\title{Multimodal Reranking for Knowledge-Intensive \\ Visual Question Answering}

\author{Haoyang Wen\footnotemark[1] \\
  Carnegie Mellon University \\
  \texttt{hwen3@cs.cmu.edu} \\\And
  Honglei Zhuang \\
  Google \\
  \texttt{hlz@google.com} \\\And
  Hamed Zamani\footnotemark[1]\thanks{$^\ast$Work performed while at Google.} \\
  University of Massachusetts Amherst\\
  \texttt{zamani@cs.umass.edu}\\\AND
  Alexander Hauptmann \\
  Carnegie Mellon University \\
  \texttt{alex@cs.cmu.edu} \\\And
  Michael Bendersky \\
  Google \\
  \texttt{bemike@google.com} \\
  }
\begin{document}

\maketitle

\begin{abstract}
Knowledge-intensive visual question answering requires models to effectively use external knowledge to help answer visual questions. A typical pipeline includes a knowledge retriever and an answer generator.
However, a retriever that utilizes local information, such as an image patch, may not provide reliable question-candidate relevance scores.
Besides, the two-tower architecture also limits the relevance score modeling of a retriever to select top candidates for answer generator reasoning.
In this paper, we introduce an additional module, a multi-modal reranker, to improve the ranking quality of knowledge candidates for answer generation. Our reranking module takes multi-modal information from both candidates and questions and performs cross-item interaction for better relevance score modeling. Experiments on OK-VQA and A-OKVQA show that multi-modal reranker from distant supervision provides consistent improvements. We also find a training-testing discrepancy with reranking in answer generation, where performance improves if training knowledge candidates are similar to or noisier than those used in testing.
\end{abstract}

\section{Introduction}
\begin{figure}
    \centering
    \includegraphics[width=0.41\textwidth]{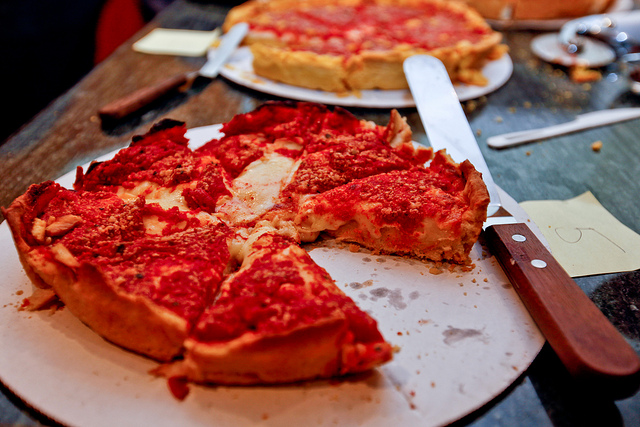}\\
    \small
    
    \begin{tabular}{l}
    \textbf{Q}: What US city is associated with this type of pizza?\\
    \textbf{A}: Chicago\\
    \end{tabular}
    \caption{An example from OK-VQA, which requires knowledge to associate deep-dish pizza and Chicago.}
    \label{fig:example}
\end{figure}

Knowledge-intensive visual question answering (KI-VQA), compared to conventional visual question answering, provides questions that cannot be directly answered with images. It requires models to use external knowledge for answer reasoning and synthesis, as shown in Figure~\ref{fig:example}.

A typical KI-VQA system contains a retrieval model to find relevant external knowledge, and an answer generator that performs reasoning over retrieved knowledge to produce the answer. One line of research investigates methods for an effective retrieval pipeline, which includes the choices of knowledge bases~\citep{DBLP:conf/mm/Li0020,garderes-etal-2020-conceptbert,luo-etal-2021-weakly}, and methods on retrieval with visual descriptions~\citep{luo-etal-2021-weakly} or image-text retrieval~\citep{gui-etal-2022-kat,DBLP:conf/nips/LinX0X0Y22}.

Answer generation models usually use retrieval relevance scores to select top candidates~\cite{gui-etal-2022-kat,DBLP:conf/nips/LinX0X0Y22}. Although achieving great success, it may sometimes provide unreliable scores, especially for retrieval using images. Because we usually split an image into a series of image patches and perform retrieval with individual patches, a high relevance score of one patch may not necessarily translate to a high overall question-candidate relevance. Besides, the two-tower architecture of a retriever model also lacks cross-item modeling for predicting precise relevance scores.

In this work, we propose to include multi-modal reranking to improve the relevance score modeling, as reranking have already shown its importance in various knowledge-intensive tasks~\citep{DBLP:journals/ftir/Liu09,lee-etal-2018-ranking,DBLP:conf/aaai/WangYGWKZCTZJ18,mao-etal-2021-reader,glass-etal-2022-re2g,DBLP:conf/sigir/HofstatterC0Z23}. The multi-modal reranking uses the multi-modal question and multi-modal knowledge items to obtain the relevance score. Specifically, we finetune a pretrained multi-modal language model~\citep{DBLP:conf/iclr/Chen0CPPSGGMB0P23} to perform a multi-modal cross-item interaction between the question and knowledge items. We train our reranker on the same dataset as answer generator training, distantly supervised by checking if answer candidates appear in the knowledge text. The benefits of this reranking component are two-folded. On one side, as other typical reranking components, it can provide more reliable relevance scores by modeling the cross-item interaction. On the other side, because most of the existing retrieval models performs uni-modal retrieval~\cite{luo-etal-2021-weakly,gui-etal-2022-kat,DBLP:conf/nips/LinX0X0Y22}, reranking with multi-modal interaction can improve the quality of retrieval by multi-modal information from question and knowledge candidates.

We perform experiments on OK-VQA~\citep{DBLP:conf/cvpr/MarinoRFM19} and A-OKQVA~\citep{DBLP:conf/eccv/SchwenkKCMM22}, based on image-text retrieval~\citep{DBLP:conf/icml/JiaYXCPPLSLD21}. The results show that the distantly-supervised reranker provides consistent improvement compared to the pipeline without a reranker. 
We also observe a training-testing discrepancy with reranking for answer generation, finding that performance improves when training knowledge candidates are similar to or noisier than testing candidates. We also find that an oracle reranker can provide a promising performance upperbound, which sheds light on future research directions in this area.

\begin{figure}
    \centering
    \includegraphics{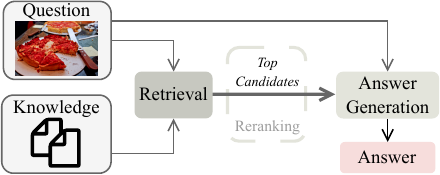}
    \caption{A basic KI-VQA framework, which first retrieves relevant top knowledge candidates with using visual question and then combine the question and retrieved knowledge candidates to generate the answer. The dashed box is our reranking module in Section~\ref{sec:reranking}.}
    \label{fig:retrieval_framework}
\end{figure}

\section{A Knowledge-Intensive Visual Question Answering Framework}
In this section, we will introduce a basic framework for KI-VQA, including image-text retrieval and answer generation, as illustrated in Figure~\ref{fig:retrieval_framework}.

\subsection{Wikipedia-Based Image Text Dataset}
In this work, we use a multi-modal knowledge base,  Wikipedia-Based Image Text Dataset (WIT)~\citep{krishna-etal-wit}. In addition to previous work that uses text from encyclopedia, WIT contains images from Wikipedia and the surrounding text at different levels, including their captions and surrounding sections. Therefore, we consider WIT as a combination of image and text knowledge.

\subsection{Image-Text Retrieval}
Previous work has explored the use of different retrieval model choices~\citep{luo-etal-2021-weakly,gui-etal-2022-kat,DBLP:conf/nips/LinX0X0Y22}. We follow one line of research that adopts image-text retrieval~\citep{gui-etal-2022-kat} using pretrained image-text language model with dual-encoder architecture~\cite{DBLP:conf/icml/RadfordKHRGASAM21,DBLP:conf/icml/JiaYXCPPLSLD21}. Following \citet{gui-etal-2022-kat}, we use sliding window with a stride to generate multiple image regions from question image. Each image region is considered as a query and will be encoded by image encoder model $\phi_i(\cdot)$. We encode captions in WIT dataset as the representation for candidates with text encoder model $\phi_t(\cdot)$, as captions in Wikipedia are generally informative. Relevance score between an image region $v_i$ and a WIT candidate $c$ is obtained with the inner product of their representations
\[
r_t(v_i, c) = \phi_i(v_i)^T\phi_t(c).
\]

\subsection{Answer Generation}
\label{sec:answer_generation}
We follow previous work~\citep{gui-etal-2022-kat,DBLP:conf/nips/LinX0X0Y22} that performs reasoning over top candidates within an encoder-decoder architecture. We also incorporate the multi-modal information~\cite{DBLP:conf/sigir/SalemiPZ23}, compared to previous work that mostly uses text-based information.

Our answer generation module is finetuned on vision language models that takes the combination of image and text as input (\textit{e.g.}, \citealp{DBLP:conf/iclr/Chen0CPPSGGMB0P23,DBLP:conf/icml/0008LSH23}).
We first encode each top candidate separately. The input of each candidate consists of question image, candidate image and text following a template\footnote{\texttt{question: <question text> candidate: <caption>}} to compose question and candidate. We encode the image with a Vision Transformer~\cite{DBLP:conf/iclr/DosovitskiyB0WZ21}, which takes a series of image patches  $\boldsymbol{x}^v=\left[x^v_1, \ldots, x^v_n\right]$, \textit{i.e.}, image tokens, to produce image representations
\[\boldsymbol{E}^v=\left[\boldsymbol{e}^v_1, \ldots, \boldsymbol{e}^v_n\right]=\enc_v(\boldsymbol{x}^v).
\]

We combine image representations and text token embeddings $\boldsymbol{E}^t$ to produce fused representations with a Transformer~\cite{DBLP:conf/nips/VaswaniSPUJGKP17}
\[
\boldsymbol{H} = \left[\boldsymbol{H}^v_q; \boldsymbol{H}^v_c; \boldsymbol{H}^t\right] = \enc_t\left(\left[\boldsymbol{E}^v_q; \boldsymbol{E}^v_c; \boldsymbol{E}^t\right]\right), %
\]
where $\boldsymbol{E}^v_q$, $\boldsymbol{H}^v_c$ represent the image token representations for question and candidate image, respectively. We also include an empty candidate that consists only of the question image and text.

While decoding, to reduce the total number of representations, we only keep the question image and text representations from the empty candidate, and the token representations that corresponds to each knowledge caption text. We concatenate these token representations to form a global representation for decoder to perform cross-attention and generate each answer token autoregressively~\cite{izacard-grave-2021-leveraging}.

\section{Multi-Modal Reranking}
\label{sec:reranking}
\begin{figure}
    \centering
    \includegraphics[width=0.48\textwidth]{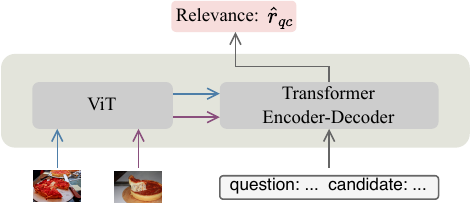}
    \caption{Framework of multimodal reranking.}
    \label{fig:reranking_framework}
\end{figure}

Vanilla retrieval-generation frameworks directly use relevance score from individual image patches. However, a high relevance from a region does not necessarily imply the overall relevance. In this section, we propose multi-modal reranking, as illustrated in Figure~\ref{fig:reranking_framework}, which takes multi-modal question and knowledge as input and produces relevance scores with cross-item interaction.

\subsection{Modeling}
Our ranking model is also finetuned on the multi-modal pretrained language model. For each question-candidate pair, we first encode the question and candidate image separately, and obtain two series of token representations $\boldsymbol{E}^v_q, \boldsymbol{E}^v_c$. Then we concatenate the two series of image token representations, with text token embeddings $\boldsymbol{E}^t$ following same template in Section~\ref{sec:answer_generation} for a Transformer to produce fused token representations
\[
\boldsymbol{H}^r = \enc_r\left(\left[\boldsymbol{E}^v_q; \boldsymbol{E}^v_c; \boldsymbol{E}^t_q; \boldsymbol{E}^t_t\right]\right).
\]
We follow \citet{DBLP:conf/sigir/Zhuang0J0MLNWB23} and use 1-step decoding to obtain the score from the unnormalized loglikelihood of a special token ``$\texttt{<extra\_id\_10>}$''
\[
\hat{r}_{qc} = \textit{Dense}\left(\textit{Dec}\left(\boldsymbol{H}^r\right)\right)_{\left(\texttt{<extra\_id\_10>}\right)}.
\]

\subsection{Ranker Training}
Because we do not have ground-truth relevance scores, we adopt distant supervision labels for reranking training. In a typical VQA setting, each answer consists of 10 answer candidate annotations. We count the number of answer candidates that occur in the knowledge candidate text as $o$. The distantly supervised relevance score is obtained similar to VQA accuracy~\cite{DBLP:conf/iccv/AntolALMBZP15}
\[
r_{qc} = \min \left\{ o/3, 1\right\}.
\]

On OK-VQA, we split the training dataset of original dataset into sub-training and -development sets. At each training step, for a question $q$, we uniformly sample a candidate set $\mathcal{C}$ from the retrieval results, and apply pairwise logistics ranking loss~\cite{DBLP:conf/icml/BurgesSRLDHH05}, which compares the ranking between all pairs of candidates in the set
\[
\ell (q) = \sum_{c\in\mathcal{C}}\sum_{c'\in\mathcal{C}}\mathbb{I}_{r_{qc} > r_{qc'}} \log \left(1 + e^{\hat{r}_{qc'}-r_{qc}}\right).
\]

\subsection{Discrepancy on Applying Reranking for Answer Generation}
\label{sec:ranker-training}
During answer generation training, it is straightforward to apply the ranking model and use the reranked top candidates as input. However, directly applying reranking on both training and testing will instead hurt the model performance. 
This is because applying the ranker on the training set, from which the ranker is trained, performs much better than when applied to the unseen test set.
As we will illustrate in Section~\ref{sec:discrepancy}, learning answer generation with higher quality ranking results while testing on lower quality ranking results will in general have a negative impact to answer generation performance. Therefore, we will keep the initial retrieval results for answer generation training, while using the reranked results for model testing.

\section{Experiments}
\subsection{Setup}
We conduct experiments on OK-VQA~\cite{DBLP:conf/cvpr/MarinoRFM19} and A-OKVQA~\cite{DBLP:conf/eccv/SchwenkKCMM22}. OK-VQA introduces visual questions that requires external knowledge. A-OKVQA further emphasizes commonsense reasoning over world knowledge. For both datasets, we evaluate the performance on the validation set. Following the standard setting, we use VQA accuracy as the metric. We use ALIGN~\cite{DBLP:conf/icml/RadfordKHRGASAM21} for image-text retrieval, and use PaLI~\cite{DBLP:conf/iclr/Chen0CPPSGGMB0P23} to initialize (vision and text) Transformers in answer generation and reranking independently. Besides retrieved knowledge candidates, we also follow REVIVE~\cite{DBLP:conf/nips/LinX0X0Y22} and use candidates generated from GPT-3. For our model with REVIVE GPT-3, we replace the last 5 candidates of the aggregated candidates with top-5 GPT-3 generated candidates from \citet{DBLP:conf/nips/LinX0X0Y22}. Appendix~\ref{app:setup} includes a detailed experimental setup.

\begin{table}[t]
    \small
    \centering
    \begin{tabular}{lc}
    \toprule
       \textbf{Methods} & \textbf{VQA Accuracy}  \\
     \midrule
      BAN+KG~\citep{DBLP:conf/mm/Li0020} & 26.7 \\
      Mucko~\citep{DBLP:conf/ijcai/ZhuYWS0W20} & 29.2 \\
      ConceptBERT~\citep{garderes-etal-2020-conceptbert} & 33.7 \\
      KRISP~\citep{DBLP:conf/cvpr/MarinoCP0R21} & 38.9 \\
      Vis-DPR~\citep{luo-etal-2021-weakly} & 39.2 \\
      MAVEx~\citep{DBLP:conf/aaai/WuLSM22} & 40.3 \\
      KAT~\citep{gui-etal-2022-kat} & 44.3 \\
      TRiG~\citep{DBLP:conf/cvpr/0013PTRWN22} & 49.4 \\
      Our model & \textbf{52.6} \\
     \hdashline
     \multicolumn{2}{l}{\textit{models with GPT-3 generated candidates}} \\
     PICa~\citep{DBLP:conf/aaai/YangGW0L0W22} & 48.0 \\
     KAT~\citep{gui-etal-2022-kat} & 53.1 \\
     REVIVE~\citep{DBLP:conf/nips/LinX0X0Y22} & 56.6 \\
     Our model + REVIVE GPT-3 & \textbf{57.2} \\
     \hdashline
      Our model w/ oracle ranking & 64.4\\
     \bottomrule
    \end{tabular}
    \caption{Results comparison on OK-VQA dataset.}
    \label{tab:okvqa-results}
\end{table}

\begin{table}[t]
    \small
    \centering
    \begin{tabular}{lc}
    \toprule
       \textbf{Methods} & \textbf{VQA Accuracy}  \\
    \midrule
       ViLBERT~\citep{DBLP:conf/nips/LuBPL19} & 30.6 \\
       LXMERT~\citep{tan-bansal-2019-lxmert} & 30.7 \\
       KRISP~\citep{DBLP:conf/cvpr/MarinoCP0R21} & 33.7 \\
       GPV-2~\citep{DBLP:conf/eccv/KamathCGKHK22} & 48.6 \\
       Our model & \textbf{51.6} \\
     \bottomrule
    \end{tabular}
    \caption{Results comparision on A-OKVQA dataset.}
    \label{tab:aokvqa-results}
\end{table}

\subsection{Results}
Our results on Table~\ref{tab:okvqa-results} and Table~\ref{tab:aokvqa-results} illustrate the performance compared to some existing work. Results on Table~\ref{tab:okvqa-results} show that we provide competitive performance compared to these systems. We also include a comparison for models with GPT-3~\cite{DBLP:conf/nips/BrownMRSKDNSSAA20} generated candidates. We find that our framework can further improve the answer generation quality with GPT-3 generated candidates from \citet{DBLP:conf/nips/LinX0X0Y22} and outperform these baselines.

We also show that an oracle ranking from distant supervision can provide a promising upper bound, indicating that there is still a large room for future work on improving ranking in this challenge.

\paragraph{Effects of Ranking Methods.}
We further conduct experiments with different ranking methods to illustrate the performance of multi-modal ranking. The results are shown in Table~\ref{tab:ranking-effects}. We compared variants of our model, including the model that generates answer directly without external knowledge, and the model with initial image retrieval without further reranking. We can find the steady improvement brought by multi-modal reranking on both datasets. We provide additional comparison to other reranking strategies in Appendix~\ref{app:reranking} and zero-shot multi-modal large models in Appendix~\ref{app:large_models} that are not instruction tuned on OK-VQA.

\subsection{Training and Testing Discrepancy}
\label{sec:discrepancy}
As we discussed in Section~\ref{sec:ranker-training}, directly applying a trained ranking model on both training and testing will hurt the performance. We further illustrate it empirically in Table~\ref{tab:discrepancy}. We can find that if the model is trained on higher quality candidates while applied on lower quality candidates, we will observe a drastic performance drop. On the contrary, when the quality in test is better than in training, we can still find steady improvement. This phenomenon indicates that an answer generator trained with higher-quality data can not effectively conduct knowledge reasoning on noisier data, and we should therefore train the model with noisier data.

\begin{table}[t]
    \small
    \centering
    \begin{tabular}{lcc}
    \toprule
        \multirow{2}{4em}{\textbf{Methods}} & \multicolumn{2}{c}{\textbf{VQA Accuracy}} \\
        & \textsc{Ok-vqa} & \textsc{A-okvqa}\\
    \midrule
       No retrieval & 50.6 & 50.4\\
       \quad + Image Retrieval & 52.1 & 50.3\\
       \quad\quad + Multimodal Reranking & \textbf{52.6} & \textbf{51.6}\\
    \bottomrule
    \end{tabular}
    \caption{Effects of multimodal reranking, compared to model without retrieval and model without reranking.}
    \label{tab:ranking-effects}
\end{table}

\begin{table}[t]
    \small
    \centering
    \begin{tabular}{llcc}
    \toprule
       \multicolumn{3}{c}{\textbf{Source of Candidates}} & \textbf{VQA} \\
       Train & Test & Discrepancy &\textbf{Accuracy}\\
    \midrule
       Retrieval & Retrieval & $\rightarrow$ & 52.1 \\
       Reranking & Reranking & $\searrow$ & 50.7 \\
       Retrieval & Reranking & $\nearrow$ & 52.6 \\
    \hdashline
       Oracle & Oracle & $\rightarrow$ & 64.4\\
       Oracle & Retrieval & $\searrow$ & 47.2\\
       Retrieval & Oracle & $\nearrow$ & 59.7\\
       Retrieval & Retrieval & $\rightarrow$ & 52.1 \\
    \bottomrule
    \end{tabular}
    \caption{Effects of discrepancy between knowledge candidates for training and testing. $\rightarrow$ means the qualities of knowledge candidates in training and test are similar. $\searrow$ means the quality in training is better than test. $\nearrow$ means the quality in test is better than training.}
    \label{tab:discrepancy}
\end{table}

\section{Related Work}
A typical knowledge-intensive visual question answering model involves a knowledge retrieval to find relevant information, and answer generator to produce the answer~\cite{DBLP:conf/mm/Li0020,garderes-etal-2020-conceptbert,luo-etal-2021-weakly,DBLP:conf/aaai/YangGW0L0W22,gui-etal-2022-kat,DBLP:conf/nips/LinX0X0Y22,DBLP:conf/sigir/SalemiPZ23,DBLP:conf/cvpr/Shao0W023}. Previous work on knowledge-intensive visual question answering explores knowledge bases in different modalities, such as text items~\cite{luo-etal-2021-weakly,gui-etal-2022-kat}, graph items~\cite{DBLP:conf/mm/Li0020,garderes-etal-2020-conceptbert}, and the composition of image items and text items~\cite{DBLP:conf/aaai/WuLSM22}. Our work differs from previous work by involving multi-modal knowledge items as the knowledge base, where each item contains both image and text information.

There is also a line of research investigating answer reranking, where they first produce a list of answer candidates, and then rerank those candidates to obtain the most reliable answer~\cite{DBLP:conf/cvpr/MarinoCP0R21,si-etal-2021-check,DBLP:conf/aaai/WuLSM22}. Instead, the focus of our work is to first retrieve a set of knowledge candidates that can help answer generation, and then improve the quality of knowledge candidate set through multimodal knowledge candidate reranking. Those selected candidates will still serves as additional knowledge input for answer generation reasoning.

\section{Conclusion}
In this paper, we introduce reranking, a critical stage for knowledge-intensive tasks, into KI-VQA. Our multi-modal reranking component takes multi-modal questions and knowledge candidates as input and perform cross-item interaction. Experiments show that our proposed multi-modal reranking can provide better knowledge candidates and improve the answer generation accuracy. Experiments on the training-testing discrepancy indicate that incorporating noisier knowledge candidates during training enhances model robustness, while training with higher quality candidates than those used in testing negatively impacts performance.

\section*{Limitations}
In this paper, we focus on applying multi-modal reranking to KI-VQA. However, because of the nature of visual data, directly adding visual information may significantly increase input size, and we will require more total memory to train the model. In this paper, to reduce the total memory use, we have a much smaller number of knowledge candidates for reasoning in answer generation module compared to previous work which only uses text-based knowledge candidates. Nevertheless, it is still important to further investigate more efficient ways to incorporate visual information.

Although multi-modal reranking achieves promising performance on knowledge-intensive visual question answering, it is still an open question that whether multi-modal reranking can be used help other vision-language tasks. Besides, it is also important to develop a benchmark to evaluate multi-modal reranking models systematically, which is not covered by this work.

Similarly, in this work, we only use ALIGN and PaLI as the pretrained model for retrieval, reranking and answer generation. Although it is natural to extend the framework in this work to other pretrained models, it is still interesting to see how it contributes to different (large and small) models. We provide some preliminary results comparing our reranking pipeline with zero-shot multi-modal large models~\cite{DBLP:conf/nips/AlayracDLMBHLMM22,DBLP:conf/icml/0008LSH23} in Appendix~\ref{app:large_models}, but we also notice that some work~\cite{DBLP:journals/corr/abs-2310-03744,DBLP:journals/corr/abs-2310-09478} uses OK-VQA as instruction tuning data, making it hard to compare/be adopted directly.

We also notice that there is another line of research investigating how to effectively use large language models for knowledge-intensive visual question answering~\cite{DBLP:conf/aaai/YangGW0L0W22,gui-etal-2022-kat,DBLP:conf/nips/LinX0X0Y22,DBLP:conf/sigir/SalemiPZ23,DBLP:conf/cvpr/Shao0W023}. Although our preliminary results show that our framework can still provide additional improvements over same the large language model queries as in \citet{DBLP:conf/nips/LinX0X0Y22}, it is still an open question to effectively use and combine the retrieval pipeline and large language model queries.
\section*{Acknowledgment}
This work was supported in part by the Google Visiting Scholar program and the Center for Intelligent Information Retrieval. Any opinions, findings and conclusions or recommendations expressed in this material are those of the authors and do not necessarily reflect those of the sponsor.

\bibliography{anthology,custom}
\appendix
\section{Experiment Setup}
\label{app:setup}
We initialize image-text retrieval module with pretrained ALIGN checkpoint, and we initialize both answer generation and multi-modal reranking module with pretrained PaLI-3b checkpoint.

In the retrieval module, we crop a question image into a series of patches with kernel size 224 with stride 64. We use each image patch to retrieve top-20 candidates and then aggregate candidates from one question image. If there are candidates that are retrieved by multiple image patches in the same image, we will keep the one with highest relevance score. We use aggregated top-20 candidates as candidates set for answer generation training and testing.

For OK-VQA, the multi-modal reranker takes 8500 of examples from training set for training, and the rest of them for model development. For each question, the reranker takes aggregated candidates from top-20 image patch retrieval as the candidate set. At each training step, we will sample 20 candidates for each question and perform pairwise logistics training. We select the reranker checkpoint based on Hits@k. The reranker is then applied to the aggregated image retrieval results to obtain the reranked relevance scores.

Answer generation is trained with batch size as 32 for 10K. Reranker is trained with batch size as 8 for 20K steps. The learning rate is 1e-4. We implement the models based on T5X~\cite{roberts2022t5x}.

\section{Additional Comparison with Other Ranking Strategies}
\label{app:reranking}
\begin{table}[htbp]
    \centering
    \small
    \begin{tabular}{lc}
    \toprule
        \textbf{Ranking Methods} & \textbf{VQA Acc.} \\
    \midrule
      Distillation~\citep{DBLP:conf/iclr/IzacardG21} & 51.5 \\
      RankT5~\cite{DBLP:conf/sigir/Zhuang0J0MLNWB23} & 52.3 \\
      Reranking & \textbf{52.6} \\
    \bottomrule
    \end{tabular}
    \caption{Effects of multimodal ranking. We can find that learning reranker using distillation from answer generator can instead hurt the performance. Our multimodal reranker trained with small data provides competitive performance even compare to RankT5 which is pretrained on large amount of data.}
    \label{tab:ranking-effects-appendix}
\end{table}
We also compare our model to the same multi-modal reranking model architecture trained with knowledge distillation from answer generation~\citep{DBLP:conf/iclr/IzacardG21} and RankT5~\citep{DBLP:conf/sigir/Zhuang0J0MLNWB23} in Table~\ref{tab:ranking-effects-appendix}. We can find that supervision from knowledge distillation can not provide reliable labels to train a reasonable reranking module. While both text-based reranking and multi-modal reranking can contribute to the performance, and multi-modal reranking can provide better performance. Especially, compared to RankT5 which is pretrained with over 500K items, our reranker only trained with around 8000 items. But it can still achieve competitive performance.

\begin{table}[htbp]
    \centering
    \small
    \begin{tabular}{lc}
    \toprule
        \textbf{Methods} & \textbf{VQA Acc.} \\
    \midrule
      BLIP-2~\citep{DBLP:conf/icml/0008LSH23} & 45.9 \\
      Flamingo-80b~\cite{DBLP:conf/nips/AlayracDLMBHLMM22} & 50.6 \\
      Our model & \textbf{52.6} \\
    \bottomrule
    \end{tabular}
    \caption{Comparison between multi-modal large models on OK-VQA datasets. We can find that our model provides promising performance compared to the zero-shot performance of those multi-modal large models.}
    \label{tab:zero-shot-llm}
\end{table}

\section{Comparison With Zero-Shot Multi-Modal Large Models}
\label{app:large_models}
We also provide additional comparison in Table~\ref{tab:zero-shot-llm} between some multi-modal large models on OK-VQA, including Flamingo-80b~\cite{DBLP:conf/nips/AlayracDLMBHLMM22} and BLIP-2~\cite{DBLP:conf/icml/0008LSH23}. We report their zero-shot performance compared to our model. The results show that smaller model can still achieve competitive performance when comparing to the zero-shot capability of those large models. We also note that there are some other multi-modal large models such as LLAVA 1.5~\cite{DBLP:journals/corr/abs-2310-03744}, MiniGPT4-V2~\cite{DBLP:journals/corr/abs-2310-09478}, which are instruction tuned with OK-VQA and therefore cannot be directly compared. But in general, our proposed framework can be extended to other multi-modal language models that take the combination of image and text input.

\end{document}